\documentclass[10pt,twocolumn,letterpaper]{article}

\usepackage[pagenumbers]{cvpr} 

\newcommand{\topic}[1]
{
\vspace{1mm}\noindent\textbf{#1}
}

\usepackage{soul} 
\usepackage{multirow}
\usepackage{amsmath, amssymb}
\usepackage{pifont}
\newcommand{\xmark}{\ding{55}}%
\usepackage{enumitem, array}
\newcolumntype{C}[1]{>{\centering\arraybackslash}p{#1}}
\newcolumntype{L}[1]{>{\raggedright\arraybackslash}p{#1}}
\newcolumntype{R}[1]{>{\raggedleft\arraybackslash}m{#1}}

\usepackage{blindtext}
\usepackage{graphicx}
\usepackage{caption}
\usepackage{adjustbox} 
\usepackage{multirow} 

\usepackage[symbol]{footmisc}

\def\m{m}  
\def\tm{\mathcal{M}}  
\def\vid{\mathcal{I}}  
\def\vidbg{\mathcal{I}_{\mathrm{bg}}}  
\def\rgba{\mathcal{O}}  

\def\a{\mathcal{\alpha}}  
\def\vidfgi{\mathcal{I}_{i,\mathrm{fg}}}
\def\err{\Delta}
\def\comp{Comp}
\def\loss{\mathcal{L}}
\def\ssr{\mathrm{SSR}}

\def\res{r}   
\def\attn{\mathbf{W}}

\definecolor{cvprblue}{rgb}{0.21,0.49,0.74}
\usepackage[pagebackref,breaklinks,colorlinks,allcolors=cvprblue]{hyperref}

\def\paperID{4994} 
\def\confName{CVPR}
\def\confYear{2025}

\title{Generative Omnimatte: Learning to Decompose Video into Layers}

\author{
Yao-Chih Lee$^{1,2,*}$ \hspace{0.9cm}
Erika Lu$^{1}$ \hspace{0.9cm}
Sarah Rumbley$^{1}$ \hspace{0.9cm}
Michal Geyer$^{1,3}$ \\
Jia-Bin Huang$^{2}$ \hspace{0.9cm}
Tali Dekel$^{1,3}$ \hspace{0.9cm}
Forrester Cole$^{1}$
\vspace{0.2cm} \\
$^1$Google DeepMind \hspace{0.8cm}
$^2$University of Maryland College Park \hspace{0.8cm}
$^3$Weizmann Institute of Science 
\vspace{0.2cm} \\
\url{https://gen-omnimatte.github.io}
\vspace{-0.6cm}
}

\begin{document}
\twocolumn[{
\renewcommand\twocolumn[1][]{#1}
\maketitle
\begin{center}
    \centering
    \captionsetup{type=figure}
    \includegraphics[trim={0 13.7cm 6.6cm 0.1cm},clip,width=\linewidth]{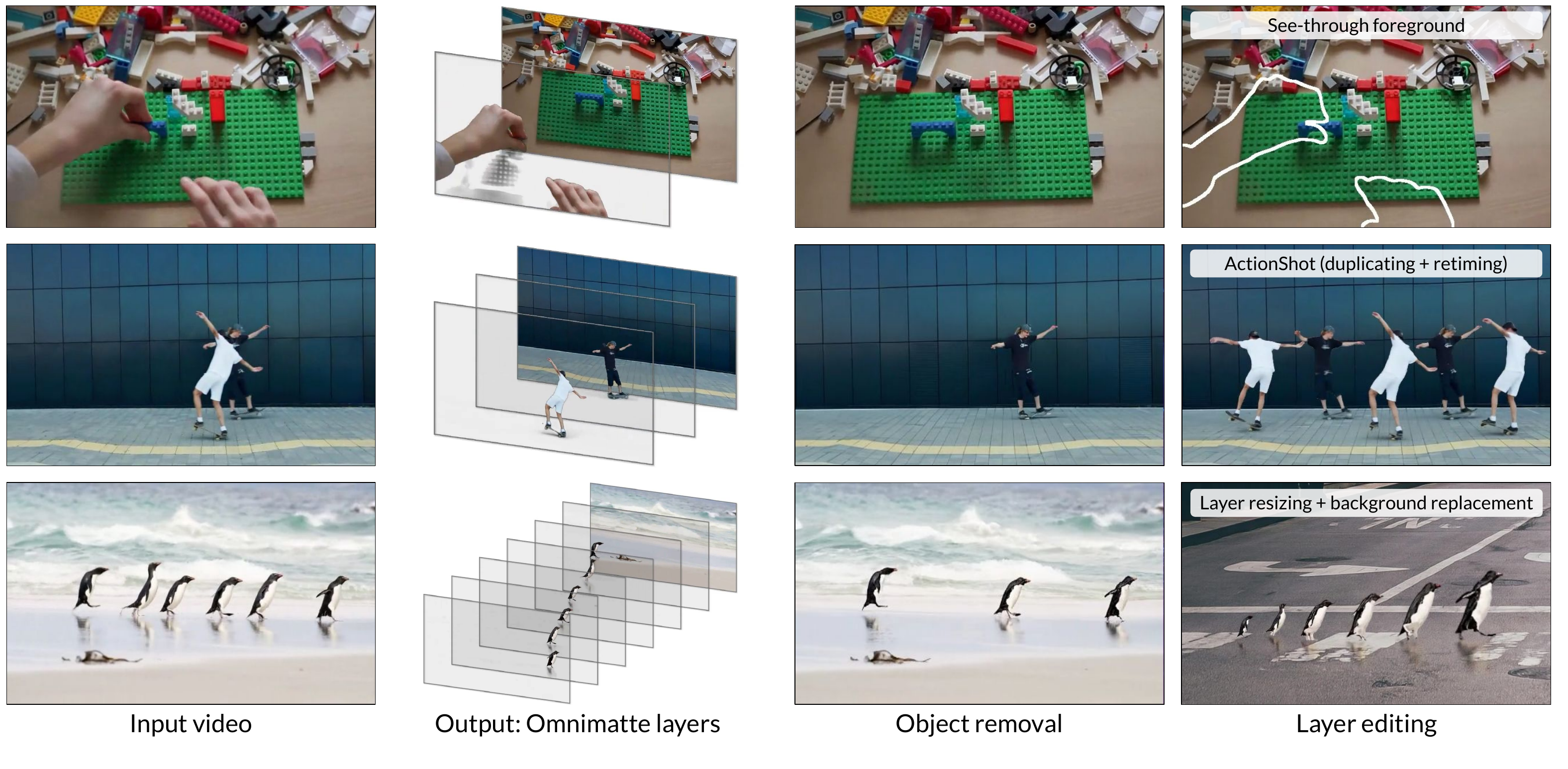}
    
    \vspace{-0.5cm}
    \captionof{figure}{
    \textbf{Generative Omnimatte.}
    Our method decomposes a video into a set of RGBA omnimatte layers, where each layer consists of a fully-visible object and its associated effects like shadows and reflections. 
    We improve upon existing work by adding a generative video prior, allowing our method to complete occluded regions (top, middle) and handle dynamic backgrounds (bottom). 
    }
    \label{fig:teaser}
    \vspace{0.2cm}
\end{center}
}]

\begin{abstract}

Given a video and a set of input object masks, an \emph{omnimatte} method aims to decompose the video into semantically meaningful layers containing individual objects along with their associated effects, such as shadows and reflections.
Existing omnimatte methods assume a static background or accurate pose and depth estimation and produce poor decompositions when these assumptions are violated. 
Furthermore, due to the lack of generative prior on natural videos, existing methods cannot complete dynamic occluded regions.
We present a novel \emph{generative} layered video decomposition framework to address the omnimatte problem. 
Our method does not assume a stationary scene or require camera pose or depth information and produces clean, complete layers, including convincing completions of occluded dynamic regions. 
Our core idea is to train a video diffusion model to identify and remove scene effects caused by a specific object. 
We show that this model can be finetuned from an existing video inpainting model with a small, carefully curated dataset, and
demonstrate high-quality decompositions and editing results for a wide range of casually captured videos containing soft shadows, glossy reflections, splashing water, and more.
\footnotetext{*Work done while Yao-Chih was an intern at Google DeepMind.}
\end{abstract}

\section{Introduction}
\label{sec:intro}
Natural videos typically intertwine complex signals about our dynamic world, such as camera motion, articulated object movement, intricate scene effects, and time-varying interaction between objects. 
Mid-level representations that model this complexity are vital for intuitive interaction and editing of video content. 
One such representation is a decomposition of video into a set of semantically meaningful, semi-transparent \emph{layers}, typically with one layer for each foreground object and the background. 
The utility of layers in creating visual effects has fueled interest in generating them from casually-captured video.

\begin{figure}
\centering
\includegraphics[trim={0.2cm 14cm 28cm 0},clip,width=\linewidth]{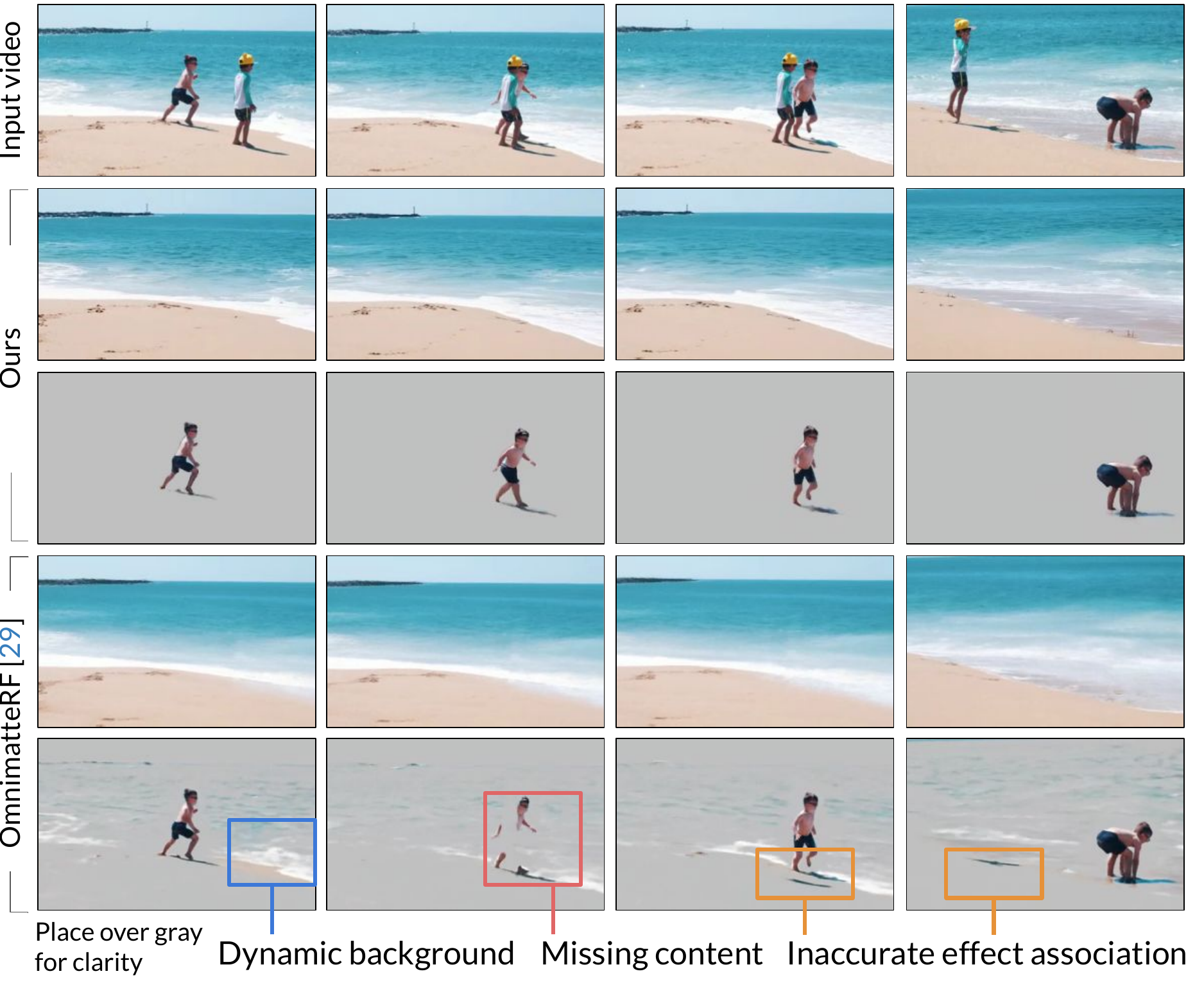}
\vspace{-0.6cm}
\caption{\textbf{Limitations of existing Omnimatte methods.}
Omnimatte methods~\cite{omnimatte,omnimatte3d,omnimatterf,factormatte} rely on restrictive motion assumptions, such as stationary background, resulting in dynamic background elements becoming entangled with foreground object layers. 
Furthermore, these methods lack a generative and semantic prior for completing occluded pixels and accurately associating effects with their corresponding objects.
}
\label{fig:limits_prev_omnimatte}
\end{figure}

\emph{Omnimatte} \cite{omnimatte,layeredneuralrendering,omnimatterf,omnimatte3d,factormatte,omnimattesp} methods produce semantically meaningful layers that contain individual objects along with their associated effects, such as shadows and reflections. 
To infer the complex space-time correlations between objects and their effects, existing methods rely heavily on restrictive assumptions such as a stationary background or accurate camera and depth estimation.
These methods fail when these assumptions are violated. 
Furthermore, existing methods cannot complete occluded regions, limiting their potential applications (Fig.~\ref{fig:limits_prev_omnimatte}).

In this work, we present a novel \emph{generative} layered video decomposition framework that overcomes previous limitations by leveraging the strong generative prior of a pretrained video diffusion model~\cite{lumiere}.
The video prior serves two purposes. First, the model's internal features reveal connections between objects and their effects (Fig.~\ref{fig:attention}) that can be exposed through finetuning, similar to recent works that apply the internal features of video diffusion models for analysis tasks~\cite{chronodepth,depthcrafter}. 
Second, 
the model can directly complete occluded areas in the layer decomposition, \emph{including dynamic regions}. 
This prior is a key factor that enables our method to dispense with the ad-hoc, limited priors of previous works~\cite{omnimatte, factormatte,omnimatte3d,omnimatterf} and the restrictions they place on the input video.

\begin{figure}
\centering
\includegraphics[trim={0.2cm 23cm 19cm 0},clip,width=\linewidth]{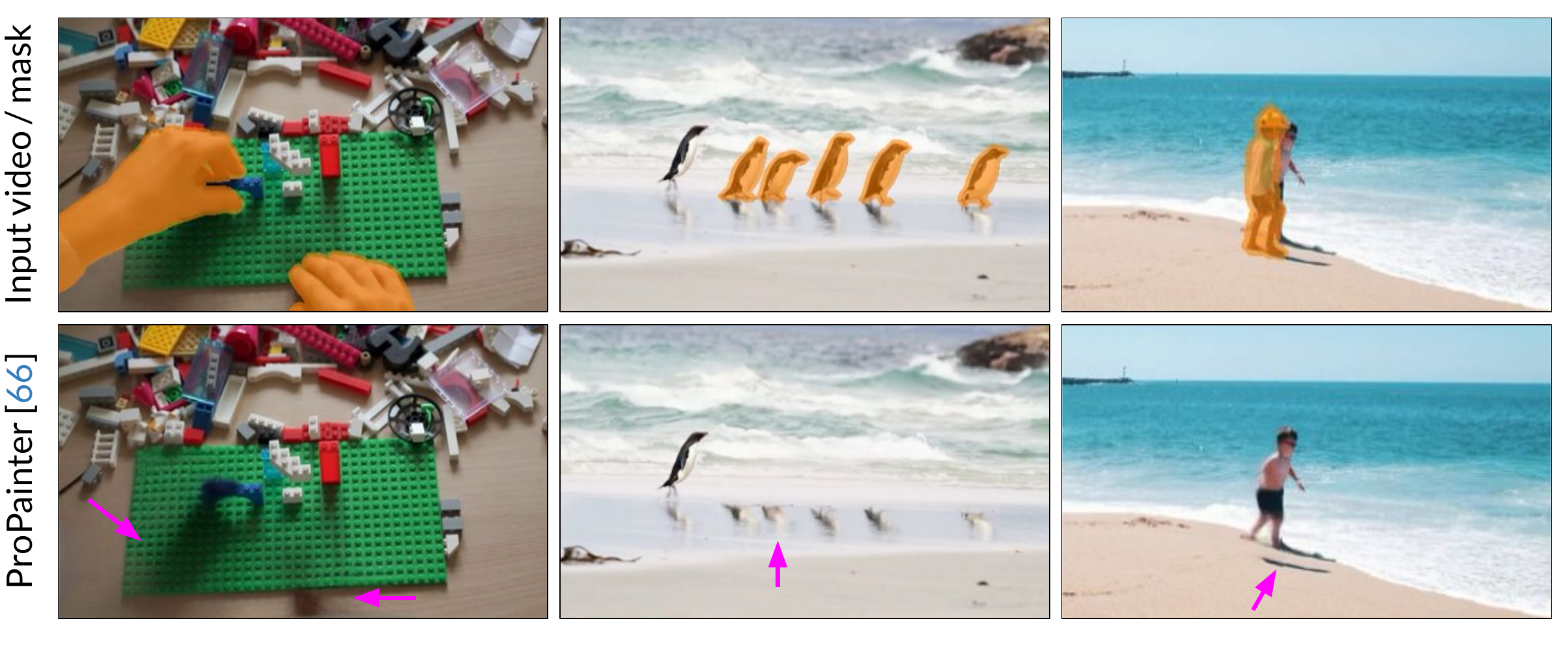}  
\vspace{-0.8cm}
\caption{
\textbf{Limitations of inpainting models for object removal.} 
 While video inpainting models (\eg,~\cite{propainter}) can complete plausible background pixels in the input mask region, they preserve the removed objects' shadows and reflections outside the mask. 
}
\label{fig:inpainting}
\end{figure}

However, existing video inpainting models are not directly applicable to the omnimatte task --- because they inpaint only the region indicated by a binary mask, they lack the flexibility to remove associated effects outside of the mask region (Fig.~\ref{fig:inpainting}). 
Instead, we finetune the video inpainting model to create an \emph{object-effect-removal} model, and apply it as a core component of a layer-decomposition system (Fig.~\ref{fig:system}). 
The model produces a ``clean-plate'' RGB background layer and initialization for the foreground layers. The final RGBA foreground layers are optimized to reconstruct the input video while remaining sparse. Since the pretrained model's features already connect objects and their effects, a small, carefully curated dataset of real and synthetic examples of object-effect associations is sufficient for finetuning.

Our method advances video layer decomposition in two directions: (i) it operates on casual videos where previous methods fail, without relying on a stationary background or camera poses, and (ii) it generates realistic, dynamic layers for both fully occluded regions and regions partially occluded by shadows or reflections. These capabilities unlock various creative editing tasks such as object removal, movement retiming, and foreground stylization (Fig.~\ref{fig:teaser}). We will release the dataset of real and synthetic layered videos used to finetune the video inpainting model.


\section{Related Work}
\label{sec:related_work}

\topic{Transparent Images and Matting.}
Transparent (RGBA) images and videos are widely used in various applications, including rotoscoping and visual effects synthesis. 
The \textit{matting} problem~\cite{porter1984compositing,wang2008mattingsurvey} aims to extract foreground RGBAs and background layers from the given images or videos.
In the visual effects industry, this layer decomposition is traditionally achieved by green-screen chroma keying, requiring complex filming setups and manual post-processing~\cite{nuke}.
Deep learning methods~\cite{lin2021real,ke2022modnet,li2021p3m,wang2024matting} approach this problem by training on large-scale alpha matte groundtruth~\cite{lin2021real,li2021p3m}.
These studies focus on estimating fine-level alpha mattes for foregrounds but cannot capture the correlated effects.
Recent advancements~\cite{transparentlayerdiff} have enabled text-to-RGBA image generation with plausible effects by finetuning Stable Diffusion~\cite{stablediff} on a transparent image dataset.
Nonetheless, decomposing \emph{videos} into layers with effect association is a different problem domain and remains under-explored.

\topic{Video Layer Decomposition.}
Decomposing a video into a set of semantically meaningful layers enables a wide range of applications, 
including video editing~\cite{layeredatlas,lee2023shape,text2live,shrivastava2024video}, motion re-timing~\cite{layeredneuralrendering}, view synthesis~\cite{ldi,mpi}, game deconstruction~\cite{smirnov2021marionette}, and obstruction removal~\cite{alayrac2019visual,alayrac2019controllable,liu2020seethrough}. 
Our work focuses on synthesizing fully-visible layers of individual objects and their associated effects from a video.

\topic{Video Matting with Associated Effects.}
Omnimatte~\cite{layeredneuralrendering,omnimatte}, inspired by early flow-based methods~\cite{wang1994representing,adelson1995layered,brostow1999motion}, leverages motion cues to decompose a video into RGBA matte layers, each comprising an object and its associated effects. 
It assumes a static background, modeled by planar homographies, and groups the time-varying residual (moving objects and effects) into foreground layers through a test-time optimization step.
Recent works extend Omnimatte by enhancing the deep image priors~\cite{omnimattesp,factormatte} and relaxing the planar homography constraint to non-rigid warping~\cite{layeredneuralrendering,deformablesprites} or 3D scene representations~\cite{omnimatterf,omnimatte3d}.
However, these approaches are limited to restrictive motion assumptions (\eg, static backgrounds, accurate pose estimation) and lack generative capabilities and semantic understanding for completing occluded areas and detecting precise object-effect associations (Fig.~\ref{fig:limits_prev_omnimatte}).
Our method, in contrast, requires no motion assumption and leverages the generative and semantic priors of our object-effect-removal model to obtain a set of fully-visible layers with accurately associated effects.

\topic{Video Generation and Inpainting.}
Recent advancements in diffusion models have significantly improved the capabilities of text-to-video generation. 
These models~\cite{lumiere,svd,ge2023preserve,videocrafter2,show1,imagenvideo,genmo,emu,cogvideox,opensora,dynamicrafter} are trained on massive datasets of video-text pairs, enabling them to generate visually appealing videos from text descriptions.

Text-to-video models have been adapted for the task of inpainting (\eg,~\cite{vidpanos,cococo,zhang2024avid}), in which they synthesize content in a user-specified mask region, conditioned on the surrounding visual context.
While existing video inpainting methods~\cite{chang2019free,gao2020flow,liu2021fuseformer,li2022e2fgvi,zhang2022flow,zhang2024avid,wu2024languageinpainting} are often employed for object removal, these models cannot handle the effects that extend beyond mask boundaries, leaving undesired shadows and reflections in the output frames (Fig.~\ref{fig:inpainting}).

\topic{Object Effect Removal.}
Work~\cite{liu2022blind,chen2024learning} in video shadow removal focuses specifically on removing cast shadows, but does not aim to perform object-shadow association, object removal, or background completion.
Recent methods~\cite{objectdrop,alzayer2024magic} fine-tune image diffusion models for object and effect removal.
However, they are unsuitable for video object removal because they lack temporal coherence when run on individual frames. 
Moreover, our goal extends beyond object removal; we aim to decompose a video into an entire set of omnimattes to enable layer-based video editing tasks.

\section{Method}
\label{sec:method}
\begin{figure*}
\centering
\includegraphics[trim={0.4cm 18cm 1.6cm 0},clip,width=\linewidth]{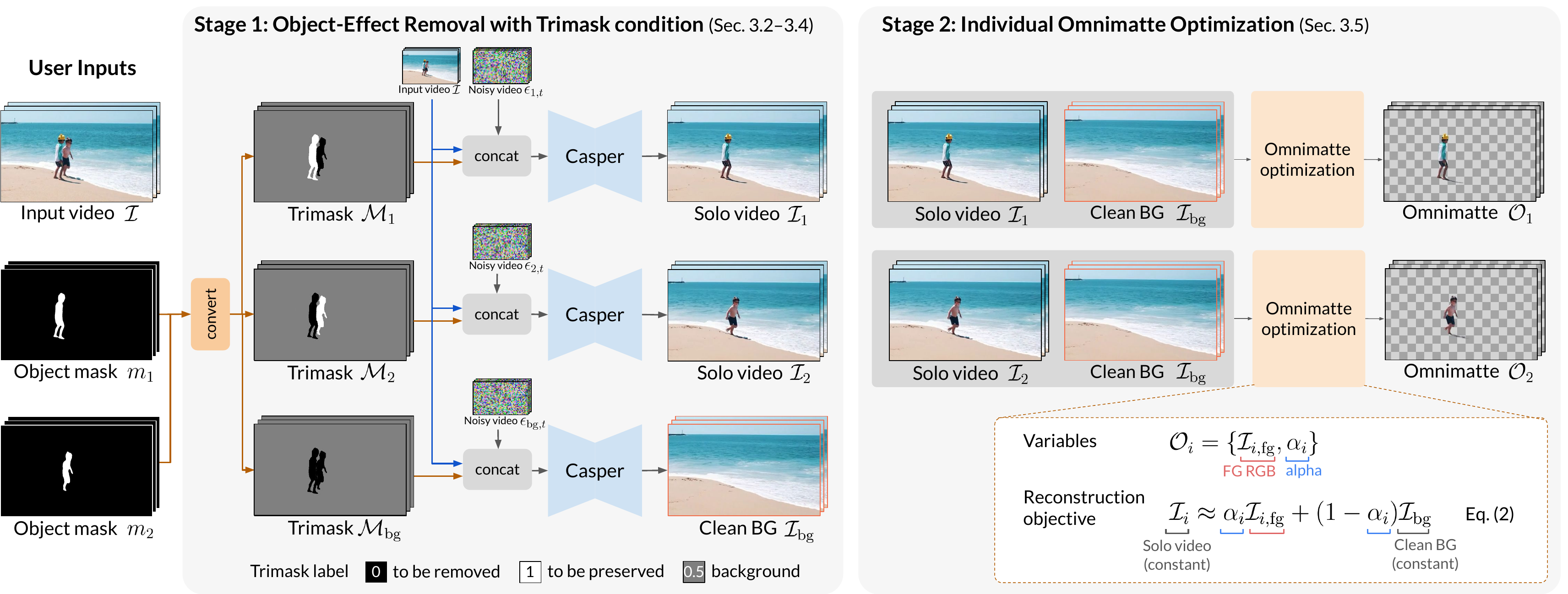}
\vspace{-0.6cm}
\caption{
\textbf{Generative omnimatte framework.} 
Given an input video and binary object masks, we first apply our object-effect-removal model, Casper, to generate a clean-plate background $\vidbg$ and a set of single-object (solo) videos $\vid_i$ applying different trimask conditions.
The trimasks specify regions to preserve (white), remove (black), and regions that potentially contain uncertain object effects (gray).
In Stage 2, a test-time optimization reconstructs the omnimatte layers $\rgba_i$ from pairs of $\vid_i$ and $\vidbg$. 
}
\vspace{-0.3cm}
\label{fig:system}
\end{figure*}

Given an RGB video $\vid$ and $N$ object masks $\{\m_i\}_{i=1}^{N}$, our goal is to generate $N+1$ video layers: a clean-plate RGB background $\vidbg$ and $N$ foreground RGBA omnimattes $\{\rgba_i\}_{i=1}^N$, each containing one object and its associated effects (such as shadows or reflections).  

Since ground-truth layered video data is scarce, we aim to apply an existing video diffusion model (Sec.~\ref{sec:base_model}) with minimal modifications. 
Large, pretrained video models have already learned the required associations between objects and their effects as part of their generative task (Sec.~\ref{sec:attention}), so we aim to bring out this ability by finetuning the model with a small dataset of layered videos (Sec.~\ref{sec:train_data}).  

The diffusion model produces RGB (without alpha), and retraining to produce alpha maps would require a large dataset of RGBA layers~\cite{transparentlayerdiff}. To avoid changing the output space of the model, we employ a two-step process (Fig.~\ref{fig:system}): 
(1) creation of the clean-plate video $\vidbg$ and $N$ ``soloed'' or single-object videos $\vid_i$ containing only object $i$ and the background, and 
(2) reconstruction of the foreground layers $\rgba_i$ from the pairs $(\vid_i, \vidbg)$ (Fig.~\ref{fig:system}).
To construct the background $\vidbg$ and solo videos $\vid_i$, we train a new video object-effect-removal model, which we dub \emph{``Casper''}\footnote{Casper: the invisible \href{https://youtu.be/wrVk-ajlKSY?feature=shared&t=33}{Friendly Ghost}}, guided by a trimask (Sec.~\ref{sec:model}). 
We then reconstruct the layers $\rgba_i$ via test-time optimization (Sec.~\ref{sec:optimization}).

\subsection{Base Video Diffusion Model}
\label{sec:base_model}
Our object-effect-removal model, \emph{Casper}, builds upon the inpainting variant of the video diffusion model, Lumiere~\cite{lumiere}. 
Lumiere is a text-to-video generator with a two-stage cascade of pixel-based diffusion models. 
The first-stage base model generates an 80-frame video at 128$\times$128px resolution from a text prompt. 
The second stage, a spatial super-resolution (SSR) model, upsamples the output of the base model to 1024$\times$1024px resolution. 

The Lumiere inpainting model~\cite{lumiere,vidpanos}, finetuned from the original text-to-video base model, takes two additional conditions: a masked RGB video with zeroed-out inpainting regions and a binary mask video $\m$. 
The base inpainting model is followed by the same SSR to achieve high-resolution quality.

Casper is finetuned from the inpainting model to perform object-and-effect removal. While Casper keeps the same model architecture as the base inpainting model, our decomposition framework introduces novel strategies for input object masking (Sec.~\ref{sec:model}). 

\subsection{Object and Effect Removal with Trimask}
\label{sec:model}

The original Lumiere inpainting model is conditioned on a binary mask $\m$ that indicates regions to be inpainted ($m=0$) and preserved ($m=1$). 
Our object-effect-removal task introduces extra ambiguity: the ``preservation'' region may not be completely preserved, but instead can be potentially changed (\eg, to erase shadows). 

To address the ambiguity, we propose a trimask condition $\tm$, derived from the binary object masks $\{\m_i\}_{i=1}^N$. 
The trimask explicitly marks three regions: the objects to remove ($\tm=0$), objects to preserve ($\tm=1$), and background areas ($\tm=0.5$) that may contain effects to be removed or preserved (Fig.~\ref{fig:system}).
To obtain the clean-plate background video $\vidbg$, we use a background trimask $\tm_{bg}$ that marks all objects as areas to remove and the background as an area to potentially modify. 
To associate effects with their corresponding $N$ objects, we aim to produce $N$ solo (single-object) videos $\vid_i$ that contain a single object and its correlated effects.
The $N$ input trimasks $\{\tm_i\}_{i=1}^N$ mark the object $i$ as a preservation area and the remaining $N-1$ objects as removal areas.
The set of $N+1$ trimasks ($\tm_{bg}$ and $\{\tm_i\}_{i=1}^N$) are fed into Casper separately, each with the same noise initialization, to generate $N$ solo videos $\vid_i$ and the background video $\vidbg$.
We obtain the binary object masks $\m_i$ using SegmentAnything2~\cite{sam2}.

Additionally, instead of a masking the RGB values in the input video,
we follow ObjectDrop~\cite{objectdrop} and keep the RGB values of the removal regions ($\tm=0$) in the input video condition, 
allowing the model to associate the content inside the mask with the corresponding effects outside the mask.

During inference, the inputs to Casper are a text prompt describing the target removal scene,
and a concatenation of the input video, trimask, and noisy video at a 128px resolution (\eg, 224$\times$128px).
The model takes 256 DDPM~\cite{ddpm} sampling steps without classifier-free guidance for inference ($\sim$12 minutes for an 80-frame video). 
We adopt temporal multidiffusion~\cite{multidiffusion} to handle longer videos.

\subsection{Effect Association Prior in Video Generator}
\label{sec:attention}

\begin{figure}
\centering
\includegraphics[trim={0.2 29.7cm 41.7cm 0cm},clip,width=0.98\linewidth]{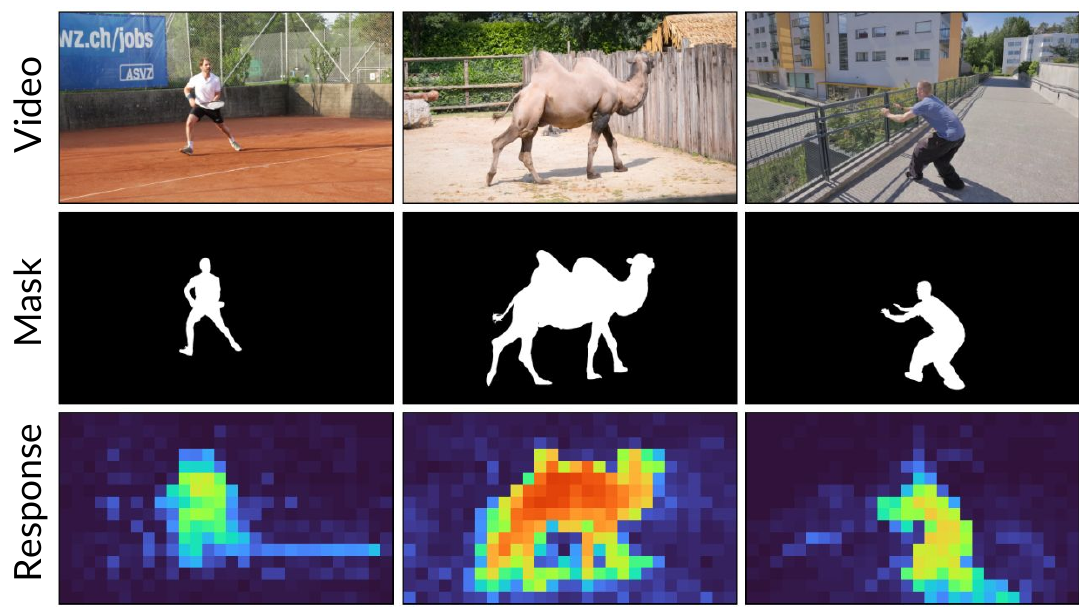}  
\vspace{-0.3cm}
\caption{
\textbf{Effect association prior in a pretrained video generation model.}
We input a video to a pretrained text-to-video generator~\cite{lumiere} using an SDEdit approach~\cite{sdedit} and analyze the spatial self-attention weights.
By measuring the attention weights between query tokens and key tokens located within target object areas, we observe that the generator can effectively associate effects with target objects. 
In this specific example, we re-noise the video to $t=0.5$ and visualize the attention in the middle bottleneck of the U-Net at the sampling step $t=0.125$. 
}
\vspace{-0.3cm}
\label{fig:attention}
\end{figure}

To investigate Lumiere's inherent understanding of object-effect associations, we analyze the self-attention pattern during denoising of a given video using SDEdit~\cite{sdedit}. 
We focus on the spatial self-attention within the frame since it is the strongest indicator of the model's prior on object-effect associations.
Specifically, we measure the self-attention weights between the query tokens and key tokens associated with the object of interest.
For a query at the spatial location $p$ in the $j^{\mathrm{th}}$ frame $\vid^j$, the response to the object is calculated as:
\begin{equation}
    \res(p) = \frac{\sum_{y\in\m^j} \attn_{p,y}}{\sum_{x\in \vid^j} \attn_{p,x}},
\end{equation}
where $\attn$ is the attention weight matrix obtained by the attention operation~\cite{transformer}, and $\attn_{p,x}$ denotes the weight between the query at $p$ and the key token at $x$.
The numerator sums the weights between the query and all the keys within object mask $\m^j$, and the denominator normalizes this sum by the total weight across the entire image space $\vid^j$. 

As visualized in Fig.~\ref{fig:attention},
we observe that the query tokens in the shadow areas exhibit attention to the object region. 
This suggests that the pretrained model~\cite{lumiere} can associate objects and their effects.
Leveraging this prior knowledge, we hypothesize that a relatively small training dataset can be sufficient to train an effective object-effect-removal model.

\subsection{Training Data for Object and Effect Removal}
\label{sec:train_data}
\begin{figure}
\centering
\includegraphics[trim={0cm 28.1cm 44cm 0cm},clip,width=\linewidth]{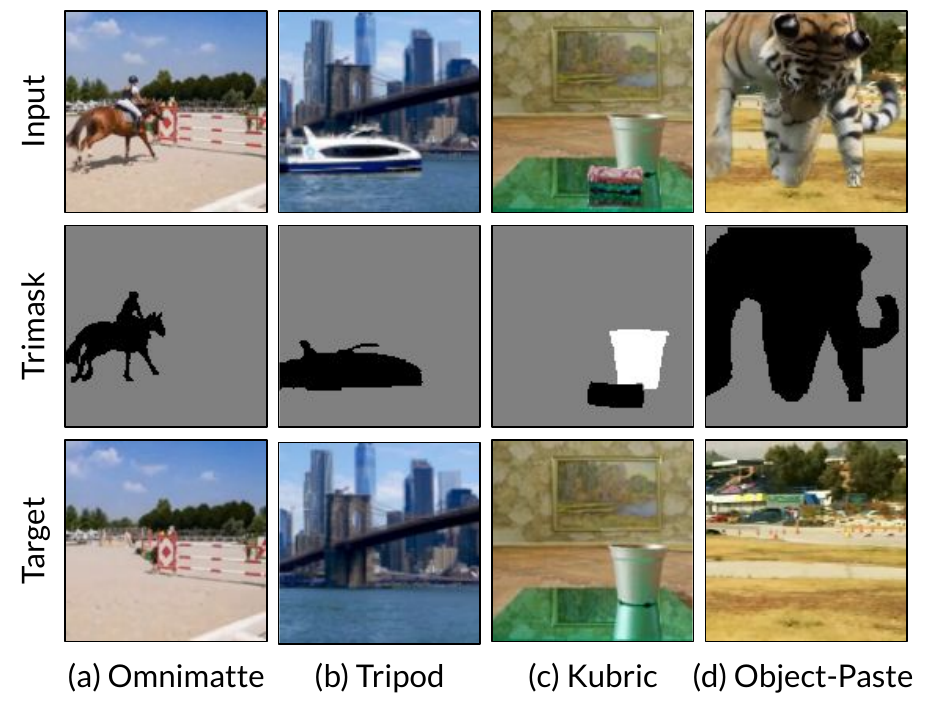}
\vspace{-0.7cm}
\caption{
\textbf{Training data for object and effect removal.}
(a) We collect omnimatte results from existing omnimatte methods to provide examples of cause-and-effect relationships in real videos.
(b) The Tripod dataset consists of videos captured with stationary cameras, providing pseudo-examples of more complex real-world scenarios, such as water effects and dynamic backgrounds.
(c) We use Kubric~\cite{kubric} to synthesize multi-object scenes with diverse reflections and shadows.
(d) We segment objects from real videos and paste them onto target real videos~\cite{vos2019}
to strengthen the model's inpainting capabilities and background preservation. 
}
\vspace{-0.2cm}
\label{fig:train_data}
\end{figure}

We curate a training dataset of real and synthetic video examples (Fig.~\ref{fig:train_data}) from the following four categories:

\noindent\textbf{Omnimatte.}
We collect 31 scenes from successful results of existing omnimatte methods~\cite{omnimatte,omnimatterf,omnimatte3d} to form the training tuples of input video, input trimask, and target background video.
These scenes, mostly from the DAVIS~\cite{davis} dataset, feature static backgrounds and single objects, providing basic examples of shadows and reflections in real-world videos.

\noindent\textbf{Tripod.}
We supplement our dataset with 15 in-the-wild videos from the web, captured by stationary cameras. 
These videos contain objects moving in and out of the scenes, which can be segmented into pseudo tuples of videos with and without the objects. 
These videos include examples of water effects (\eg, reflections, splashing, ripples) and ambient background motion. 
We augment these videos with Ken Burns effects to simulate camera motion.

\noindent\textbf{Kubric.}
While the real-world videos provide valuable object-effect examples, they primarily depict single-object scenes, limiting the model's ability to learn effect association with multiple objects.
Hence, we incorporate 569 synthetic videos generated using Kubric~\cite{kubric}.
By rendering multi-object scenes in Blender~\cite{blender} and making objects transparent, we create ground-truth removal videos that preserve the physical interactions and motion of the remaining scene elements.
Moreover, we observe that many real-world scenarios exhibit multiple instances of the same object type in a scene, such as dogs, pedestrians, or vehicles. 
Therefore, we generate scenes with duplicated objects to train the model to handle multiple similar objects. 
We also insert reflective surfaces in the scenes to simulate water and glass reflections.

\noindent\textbf{Object-Paste.}
We synthesize 1024 video tuples from real videos in the YouTube-VOS Dataset~\cite{vos2019}, using SegmentAnything2~\cite{sam2} to crop objects from a random video and paste them onto a target video. 
The training input and target are the composited video and original video, respectively. 
Training on this data strengthens the model's inpainting and background preservation abilities.

The text prompts of the training data are captioned by BLIP-2~\cite{blip2}, describing the target video that the object-effect-removal model should learn to generate. 
We augment the dataset with spatial horizontal flipping, temporal flipping, and random cropping to a 128$\times$128px resolution. 
We balance the four data categories with different ratios, using approximately equal proportions (50\%) of real and synthesized data. 
Please refer to the Supplementary Material for the training details.

\subsection{Omnimatte Optimization}
\label{sec:optimization}

As described in Sec.~\ref{sec:model}, we obtain solo videos $\{\vid_i\}_{i=1}^N$ and the background $\vidbg$ from an $N$-object video input. 
Given each pair of $(\vid_i, \vidbg)$, 
the goal is to reconstruct an omnimatte layer, $\rgba_i$, such that its composition with the background, $\vidbg$, matches the solo video, $\vid_i$:
\begin{equation}
\label{eq:composition}
    \vid_i \approx \comp(\vidbg, \vidfgi, \a_i) = \a_i \vidfgi  + (1 - \a_i) \vidbg,
\end{equation}
where $\vidfgi$ and $\a_i$ are the RGB and alpha channels of the omnimatte $\rgba_i$. 

\noindent\textbf{Variables.} 
Unlike previous methods~\cite{omnimatte,factormatte,omnimatte3d,omnimatterf} that jointly optimize the background scene model with the foreground omnimattes, our method first produces a clean background video which is kept constant during the optimization. 
Therefore, the variables to optimize are the foreground RGB $\vidfgi$ and alpha $\a_i$. 
Following existing methods~\cite{omnimatte,omnimatte3d,omnimatterf}, we use a U-Net to generate smooth alphas from the inputs $\vid_i$ and error map $\err_i=|\vid_i-\vidbg|$. We directly optimize the pixel values of the RGB $\vidfgi$ for efficiency, with a sigmoid to constrain the range to $[0, 1]$.

\noindent\textbf{Loss function.}
Similar to~\cite{omnimatte,omnimatte3d,omnimatterf}, we use a reconstruction loss, $\loss_{\mathrm{recon}}$ (the difference between the target $\vid$ and the composited video) to drive the optimization:
\begin{equation}
    \loss_{\mathrm{recon}} = \|\vid_i - \comp(\vidbg, \vidfgi, \a_i)\|_2.
\end{equation}

To prevent the foreground omnimatte from being overly sensitive to reconstruction error, 
we add $L_0$- and $L_1$-sparsity regularization on the alpha map:
\begin{equation}
    \loss_{\mathrm{sparsity}} = \beta_1 \|\a_i\|_1 + \beta_0 \Phi_0(\a_i),
\end{equation}
where $\Phi_0$ denotes an approximate $L_0$ proposed in~\cite{omnimatte}, and $\beta_1$ and $\beta_0$ are the hyper-parameters balancing the loss.

To optimize the alpha network from scratch, a mask supervision loss, $\loss_{\mathrm{mask}}=\|\m_i-\a_i\|_2$, is used to guide the learning of alpha $\a_i$ from the input object mask $\m_i$. 
The loss is gradually reduced over the optimization iterations. 
The final omnimatte $\rgba_i$ is obtained by minimizing the total loss of $\loss_{\mathrm{recon}} + \lambda_{\mathrm{sparsity}}\loss_{\mathrm{sparsity}} + \lambda_{\mathrm{mask}}\loss_{\mathrm{mask}}$. 

\noindent\textbf{Upsampling and Detail Transfer.} 
The Casper model has a resolution of 128px (\eg, 224$\times$128) inherited from the Lumiere base stage. Lumiere's SSR stage~\cite{lumiere} is used to upsample the base model outputs to the target resolution 640$\times$384. The RGBA omnimatte layers $\rgba_i$ are first optimized at the base resolution, then upsampled and optimized further at the target resolution. Finally, a detail transfer step~\cite{layeredneuralrendering,omnimatte,omnimatte3d} is applied to fully opaque regions. Please see SM for further details.

\begin{figure}
\centering
\includegraphics[trim={0.1cm 24.6cm 45cm 0cm},clip,width=\linewidth]{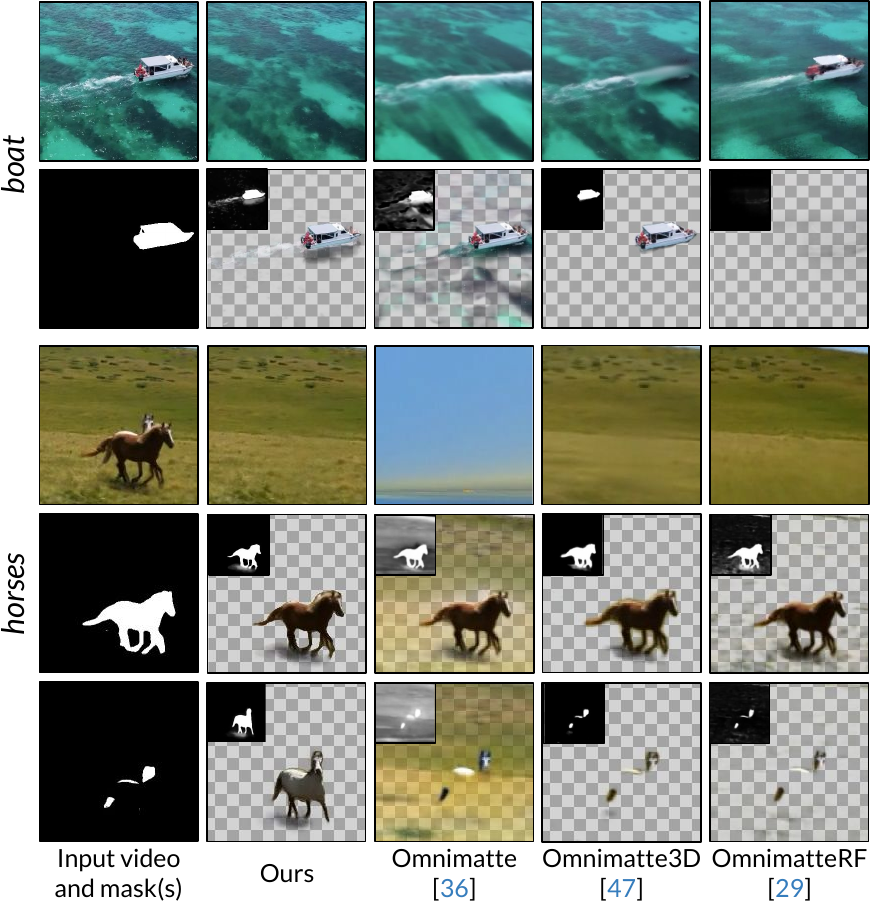}  
\vspace{-0.7cm}
\caption{
\textbf{Qualitative comparisons with omnimatte methods.}
For each example, we show the background layer and foreground omnimattes generated by different methods in the top and bottom rows, respectively.
Existing methods~\cite{omnimatte,omnimatte3d,omnimatterf} struggle with relatively stationary foregounds (``boat''), and struggle to complete heavily occluded objects (``horses'').
While~\cite{omnimatte3d,omnimatterf} can handle scenes with parallax (``horses''), the background layer appears blurry due to imperfect camera pose and depth estimation. 
}
\vspace{-0.6cm}
\label{fig:omnimatte_comparison}
\end{figure}
\begin{figure*}
\centering
\includegraphics[trim={0 20.5cm 10cm 0.5cm},clip,width=\linewidth]{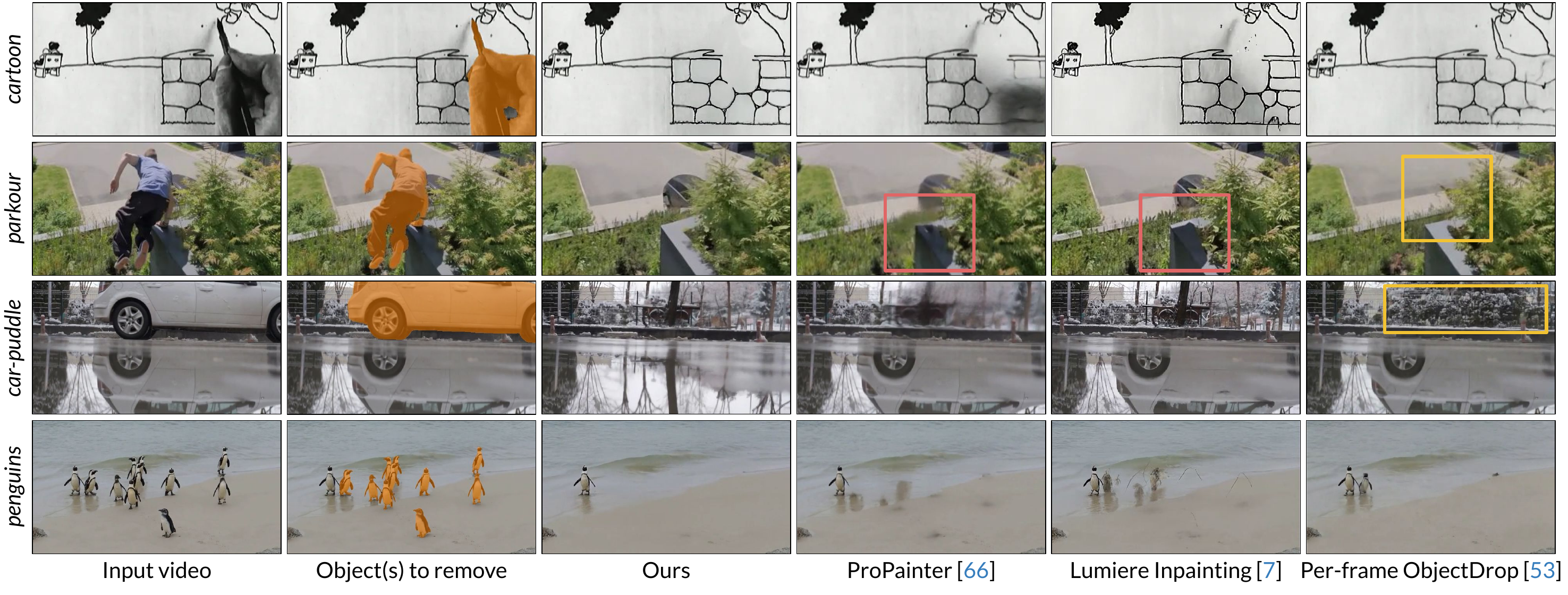}
\vspace{-0.8cm}
\caption{\textbf{Visual comparison on object removal.}
We compare our removal model with state-of-the-art video inpainting models~\cite{propainter,lumiere} and per-frame ObjectDrop~\cite{objectdrop}. 
Inpainting models fail to remove effects that extend beyond the provided mask boundaries. 
ObjectDrop processes video frames independently, lacking temporal information to preserve backgrounds and produce temporally coherent results. 
We encourage readers to view the supplementary material for video comparisons.
}
\vspace{-0.3cm}
\label{fig:removal_comparison}
\end{figure*}

\section{Results}
\label{sec:results}
We compare our method with SOTA omnimatte methods~\cite{omnimatte3d,omnimatterf}. 
We also compare Casper with video inpainting models~\cite{propainter,lumiere} and ObjectDrop~\cite{objectdrop} on object-removal tasks. 
Note that ObjectDrop is an image-based model, and thus, it processes each video frame independently.
We obtain the results of~\cite{objectdrop,omnimatte3d} from the respective authors. 

\subsection{Qualitative Results}
We compare omnimatte methods in Fig.~\ref{fig:omnimatte_comparison}. 
In ``boat'', existing approaches fail to separate the boat's wake from the background layer, while our method correctly places it in the boat's layer.
For ``horses'', Omnimatte3D~\cite{omnimatte3d} and OmnimatteRF~\cite{omnimatterf} produce blurry background layers because their 3D-aware background representations are sensitive to camera pose estimation quality. Furthermore, none of the existing Omnimatte methods can disocclude the horse in the last row.
In contrast, our method does not require camera pose information and can inpaint heavily occluded objects.

Fig.~\ref{fig:removal_comparison} compares our Casper model with existing methods for object removal. 
Video inpainting models~\cite{propainter,lumiere} fail to remove soft shadows and reflections outside the input masks.
ObjectDrop~\cite{objectdrop} can remove shadows in ``cartoon'' and ``parkour'', but it processes frames independently and inpaints regions without global context. 
This leads to inconsistent hallucinations, such as 
a missing car in ``parkour'', and inconsistent background trees in ``car-puddle''.
Our model effectively removes soft shadows (``cartoon''), handles shadows cast on complex structures (``parkour''), and excels at multi-object removal (``penguins''). 
We encourage readers to view SM for video comparisons.

\subsection{Quantitative comparison}
\begin{table}
\caption{\textbf{Quantitative comparison.}
Following the benchmark established in OmnimatteRF~\cite{omnimatterf}, we evaluate the removal quality on background videos of 10 synthetic scenes in two categories. Our method achieves the best overall scores in both PSNR and LPIPS metrics. We adopt the numbers of~\cite{omnimatte,layeredatlas,omnimatterf} reported in~\cite{omnimatterf}. Best results are highlighted in red and second-best in yellow. Results marked as ``-'' indicate failures in some scenes (\eg, all-zero outputs in~\cite{omnimatte3d}). We present the per-scene scores in SM.
}
\vspace{-0.13cm}
\resizebox{\linewidth}{!}{
\begin{tabular}{l|cccc|cc}
\toprule
Scene & \multicolumn{2}{c}{Movie} & \multicolumn{2}{c|}{Kubric} & \multicolumn{2}{c}{Average} \\
Metric & PSNR$\uparrow$ & LPIPS$\downarrow$ & PSNR$\uparrow$ & LPIPS$\downarrow$ & PSNR$\uparrow$ & LPIPS$\downarrow$ \\ \midrule
ObjectDrop~\cite{objectdrop} & 28.05 & 0.124 & 34.22 & 0.083 & 31.14 & 0.104 \\
Propainter~\cite{propainter} & 27.44 & 0.114 & 34.67 & 0.056 & 31.06 & 0.085 \\
Lumiere inpainting~\cite{lumiere} & 26.62 & 0.148 & 31.46 & 0.157 & 29.04 & 0.153 \\
Ominmatte~\cite{omnimatte} & 21.76 & 0.239 & 26.81 & 0.207 & 24.29 & 0.223 \\
LNA~\cite{layeredatlas} & 23.10 & 0.129 & - & - & - & - \\
Omnimatte3D~\cite{omnimatte3d} & - & - & - & - & - & - \\
OmnimatteRF~\cite{omnimatterf} & \cellcolor[HTML]{FFCCC9}33.86 & \cellcolor[HTML]{FFCCC9}0.017 & \cellcolor[HTML]{FFEEBF}40.91 & \cellcolor[HTML]{FFEEBF}0.028 & \cellcolor[HTML]{FFEEBF}37.38 & \cellcolor[HTML]{FFEEBF}0.023 \\
Ours & \cellcolor[HTML]{FFEEBF}32.69 & \cellcolor[HTML]{FFEEBF}0.030 & \cellcolor[HTML]{FFCCC9}44.07 & \cellcolor[HTML]{FFCCC9}0.010 & \cellcolor[HTML]{FFCCC9}38.38 & \cellcolor[HTML]{FFCCC9}0.020 \\ \bottomrule
\end{tabular}
}
\vspace{-0.3cm}
\label{tbl:quantitative_comparison}
\end{table}

We adopt the evaluation protocol from OmnimatteRF~\cite{omnimatterf} to assess background layer reconstruction on ten synthetic scenes. 
The synthetic dataset comprises 5 Movie scenes and 5 Kubric scenes generated by~\cite{d2nerf}.
Each scene has a corresponding ground-truth background without foreground objects and effects. We use PSNR and LPIPS~\cite{lpips} as evaluation metrics.
We confirm that our Kubric training data does not overlap with the Kubric scenes used in the evaluation.

The quantitative results are presented in Table~\ref{tbl:quantitative_comparison}.
Existing layer decomposition methods perform a global static scene optimization for each video.
Omnimatte~\cite{omnimatte} and Layered Neural Atlas~\cite{layeredatlas} utilize 2D motion models and thus struggle to handle parallax.
Omnimatte3D~\cite{omnimatte3d} fails to construct a background scene model in two cases and struggles with stationary foreground objects in the Movie scenes.
OmnimatteRF~\cite{omnimatterf} applies an additional background re-training step to further improve removal results and, in some cases, can complete the inpainting regions more accurately due to the global scene model optimization.
Overall, our method achieves the best performance in both metrics.
We provide visual comparisons of the evaluation scenes in SM.

\subsection{Ablation study.}
\label{sec:ablation}
\begin{figure}
\centering
\includegraphics[trim={0.2cm 20.7cm 33.5cm 0.3cm},clip,width=\linewidth]{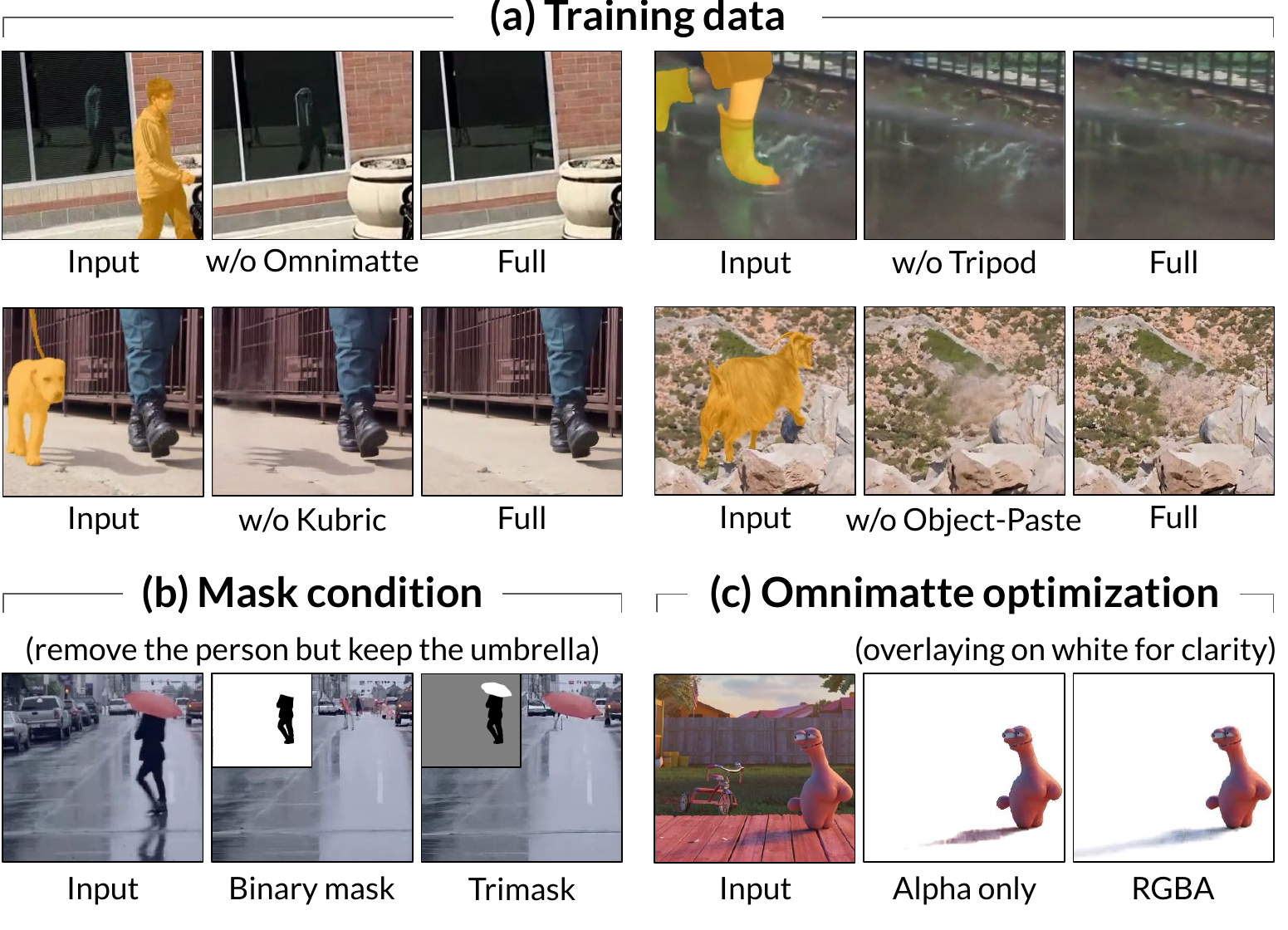}
\vspace{-0.85cm}
\caption{\textbf{Ablation study.}
(a) We ablate four different categories of training examples to validate their individual contributions. 
(b) The proposed trimask allows for greater control over defining regions to preserve or alter.
Models trained with binary masks lack this flexibility and may remove objects that are intended to be kept (\eg, the umbrella).
(c) Optimizing the foreground RGB of omnimattes rather than using pixels directly from the input video results in cleaner foreground layers. Best viewed zoomed-in on a display.
}
\vspace{-0.2cm}
\label{fig:ablation}
\end{figure}

\noindent{\textbf{Training data.}}
We assess the individual contributions of each dataset category to our model's performance in Fig.~\ref{fig:ablation}a.
The \emph{Omnimatte} data provides real-world shadow and reflection scenes. 
The \emph{Tripod} data complements \emph{Omnimatte} by providing additional real-world scenarios, including specific examples of water reflections, to better handle water effects.
Our \emph{Kubric} synthetic data strengthens the model's ability to handle 
multi-object scenes. 
\emph{Object-Paste}, synthesized by overlaying objects onto real videos, reduces undesired background changes and improves inpainting quality.

\noindent{\textbf{Mask condition.}}
As shown in Fig.~\ref{fig:ablation}b, the proposed trimask explicitly defines the regions to be removed or preserved, thereby enabling more accurate handling of multi-object scenarios.
In contrast, the model trained on binary masks is susceptible to ambiguity, potentially leading to undesired removal of objects meant to be preserved.

\noindent{\textbf{Omnimatte optimization.}}
The omnimatte reconstruction optimizes both foreground RGB and alpha channels. 
If the foreground RGB channel is not optimized and instead fixed as the input foreground video, the resulting omnimatte may exhibit increased background color contamination and hinder the generation of soft alpha mattes. (Fig.~\ref{fig:ablation}c).


\subsection{Applications}
\begin{figure}
\centering
\includegraphics[trim={0 25.5cm 28.5cm 0},clip,width=\linewidth]{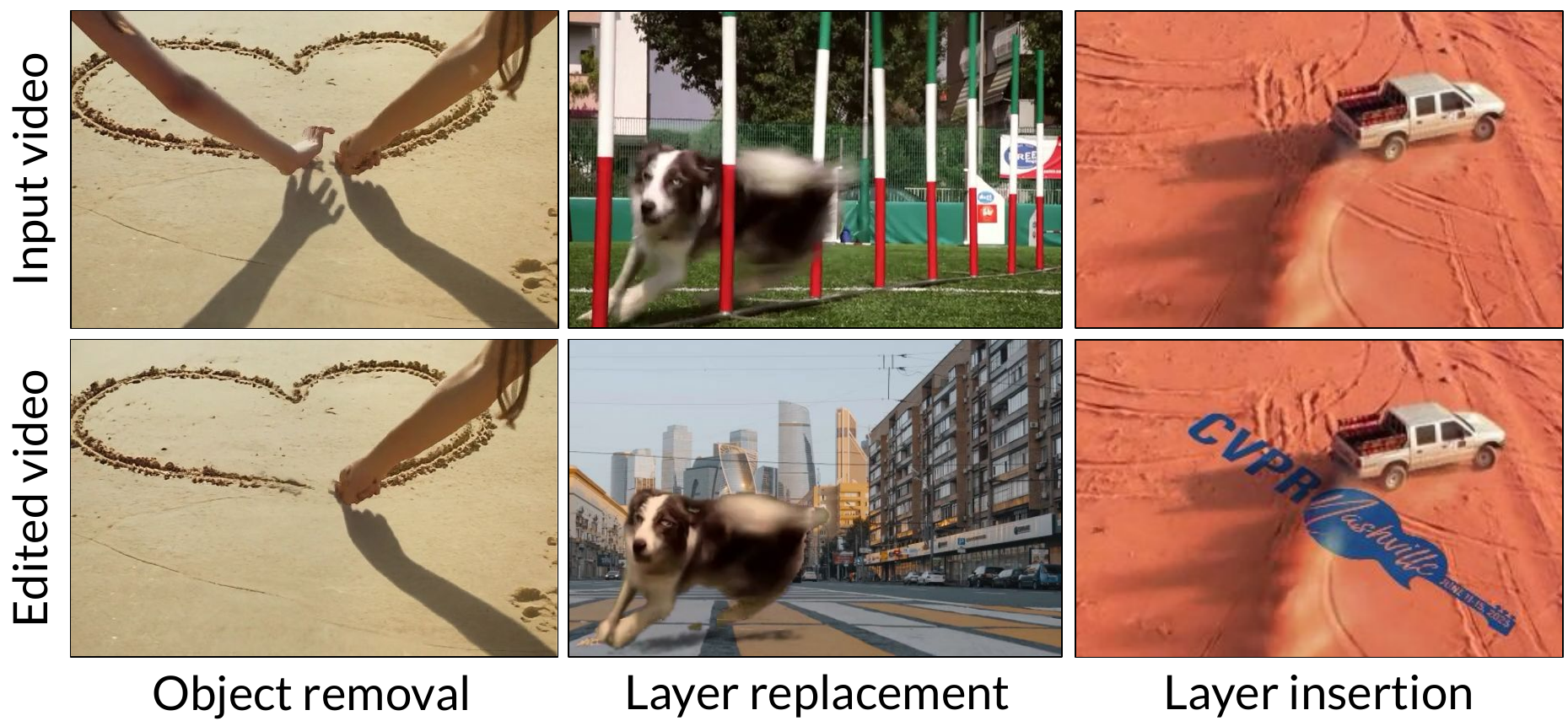}  
\vspace{-0.7cm}
\caption{\textbf{Applications.}
Our omnimattes enable diverse video editing tasks. Please refer to SM for full video demonstrations.
}
\vspace{-0.3cm}
\label{fig:apps}
\end{figure}
As showcased in Fig.~\ref{fig:teaser} and Fig.~\ref{fig:apps}, our omnimattes enable a wide range of video editing for users, such as layer insertion and replacement. 
Each layer can also be independently edited with different transformations, stylizations, and time offsets for motion retiming~\cite{layeredneuralrendering}. 
The edited layers can then be recomposed seamlessly to generate novel videos.


\section{Discussion and Limitations}
\label{sec:conclusion}

Our method overcomes the limitations of existing omnimatte approaches by incorporating a finetuned video inpainting model that performs object and effect removal. This is achieved by finetuning a pretrained inpainting model on a small training set, without modifying the architecture to generate an additional alpha channel. The RGB outputs of the video model are then used to reconstruct the RGBA omnimatte layers using an optimization step. 

\begin{figure}
\centering
\includegraphics[trim={0 24cm 22cm 0},clip,width=\linewidth]{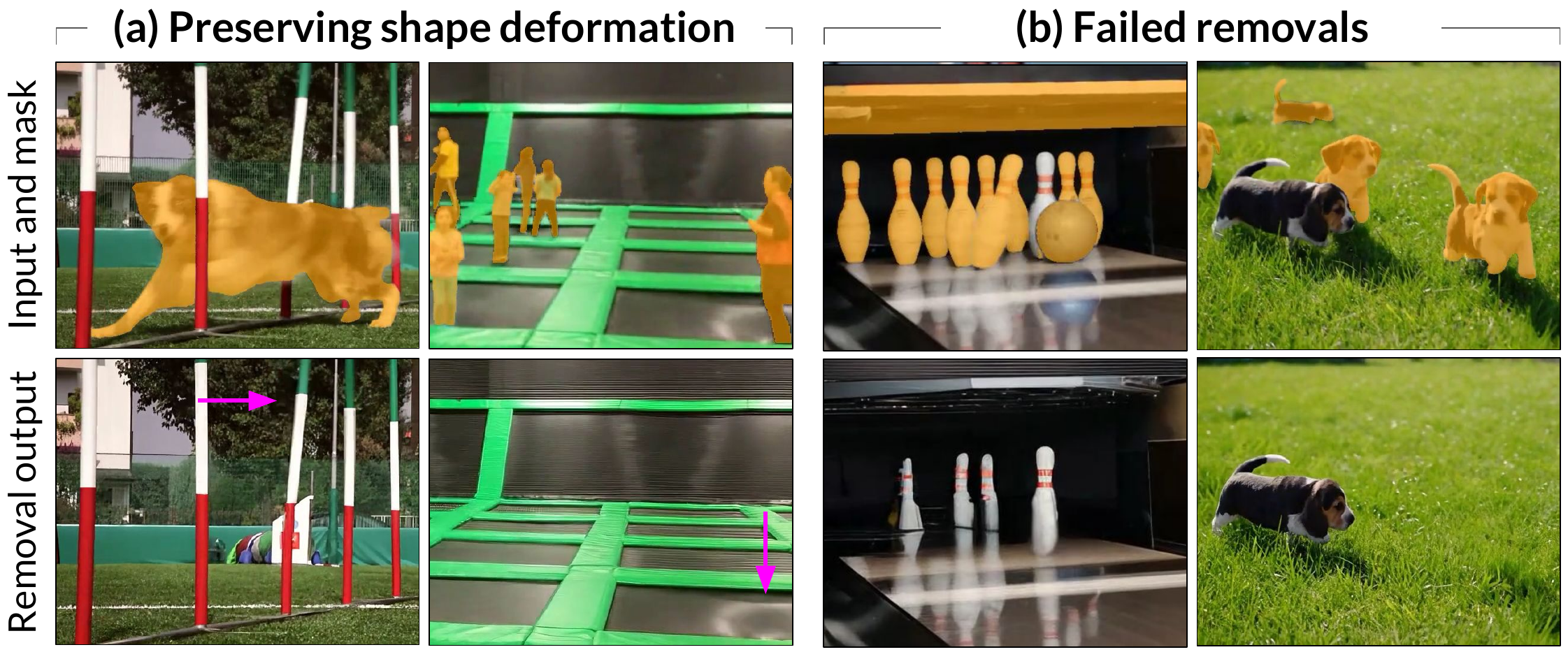}
\vspace{-0.7cm}
\caption{\textbf{Limitations.}
(a) Casper effectively removes objects and their associated 
shadows and reflections, while 
keeping shape deformation in the outputs.
(b) In challenging multi-object 
cases, Casper may struggle to accurately separate 
individual effects.
}
\vspace{-0.2cm}
\label{fig:our_fails}
\end{figure}

The finetuned model inherits the original diffusion model's learned prior on natural videos, allowing for strong generalization to scenarios not present in the training set (\eg, human hands in ``lego,'' ``cartoon''). However, this reliance on a data-driven prior can also limit the range of effects the method can handle. For example, since we did not include training data capturing physical deformations, our current model does not remove effects such as bending poles or trampolines (Fig.~\ref{fig:our_fails}a). However, our ablation experiments (Fig.~\ref{fig:ablation}) indicate the model responds well to small additions to the training data, suggesting that it may be possible capture these effects by curating additional data. 

We also observed challenging multi-object cases where, despite apparently appropriate training data, our model fails to correctly remove effects (Fig.~\ref{fig:our_fails}b). We hypothesize the model may need to be trained with additional information (\eg, instance segmentation) to disambiguate 
objects and their effects when multiple, very similar objects are present.

\begin{figure}
\centering
\includegraphics[trim={0cm 18.3cm 5.3cm 0},clip,width=\linewidth]{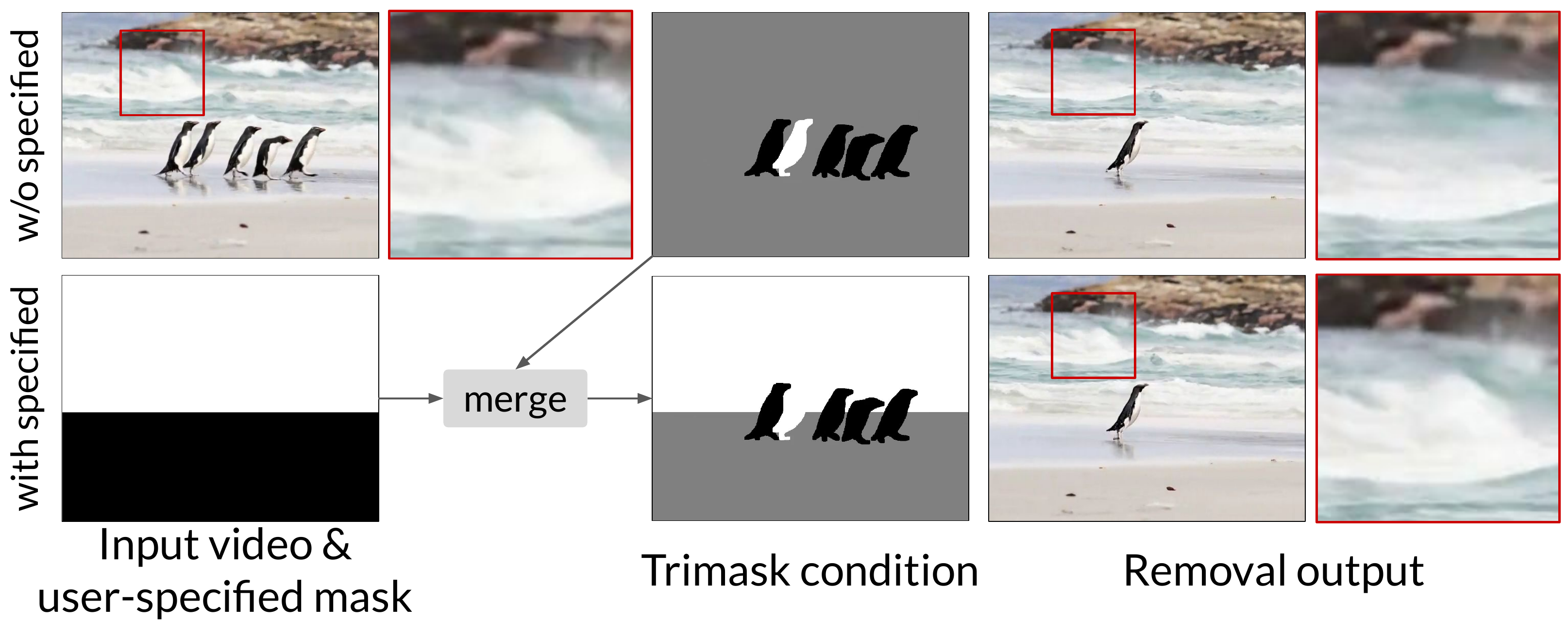}
\vspace{-0.6cm}
\caption{\textbf{User-specified trimask.}
Casper may introduce undesired changes to dynamic background elements (\eg, wave details). To mitigate this, users may specify a coarse preservation region to merge with the trimasks. The zoomed-in boxes show precise detail preservation in the specified background area.
}
\vspace{-0.3cm}
\label{fig:user_trimap}
\end{figure}

Finally, we observe some cases where Casper will associate unrelated background effects with a foreground layer, such as the waves in Fig.~\ref{fig:user_trimap}. Our system allows the user to modify the trimask to preserve background areas in simple cases (Fig.~\ref{fig:user_trimap}, mid). In future work, such lightweight user input can be used to enlarge and improve the training set. 

We present a new, generative approach to video layer decomposition. The current method already outperforms previous methods based on more limited priors, and will continue to improve as more ground-truth layered video data becomes available and generative video priors gain strength.

\topic{Acknowledgments.}
We would like to thank Charles Herrmann, Daniel Geng, Junhwa Hur, Hadi Alzayer, Deqing Sun, Daniel Vlasic, Miki Rubinstein, and Bill Freeman for their insightful discussions and generous support.
We sincerely appreciate Ray Li, Scott Wisdom, Andrei Kapishnikov, Matthew Grimes, AJ Maschinot, and Shu-Jung Han for their invaluable help in filming testing videos and contributing creative editing ideas.


{
    \small
    \bibliographystyle{ieeenat_fullname}
    \bibliography{main}
}

\clearpage

\appendix
\noindent\textbf{\Large Appendix}\vspace{0.3cm}

We provide additional discussions (Sec.~\ref{sec:supp_discussion}), as well as further details on the training process of our object-effect-removal model (Sec.~\ref{sec:supp_training}), 
the inference pipeline for object and effect removal (Sec.~\ref{sec:supp_inference}),
the optimization details for omnimatte reconstruction (Sec.~\ref{sec:supp_omptimization}),
runtime analysis (Sec.~\ref{sec:supp_runtime}),
the quantitative comparisons (Sec.~\ref{sec:supp_eval}), the quantitative ablation study (Sec.~\ref{sec:supp_ablation}).
The video comparisons, training video examples, and self-attention visualization are provided in the \href{https://gen-omnimatte.github.io}{project page}.

\section{Additional Discussions}
\label{sec:supp_discussion}
\begin{figure}
\centering
\includegraphics[trim={0cm 24cm 25cm 0.05cm},clip,width=\linewidth]{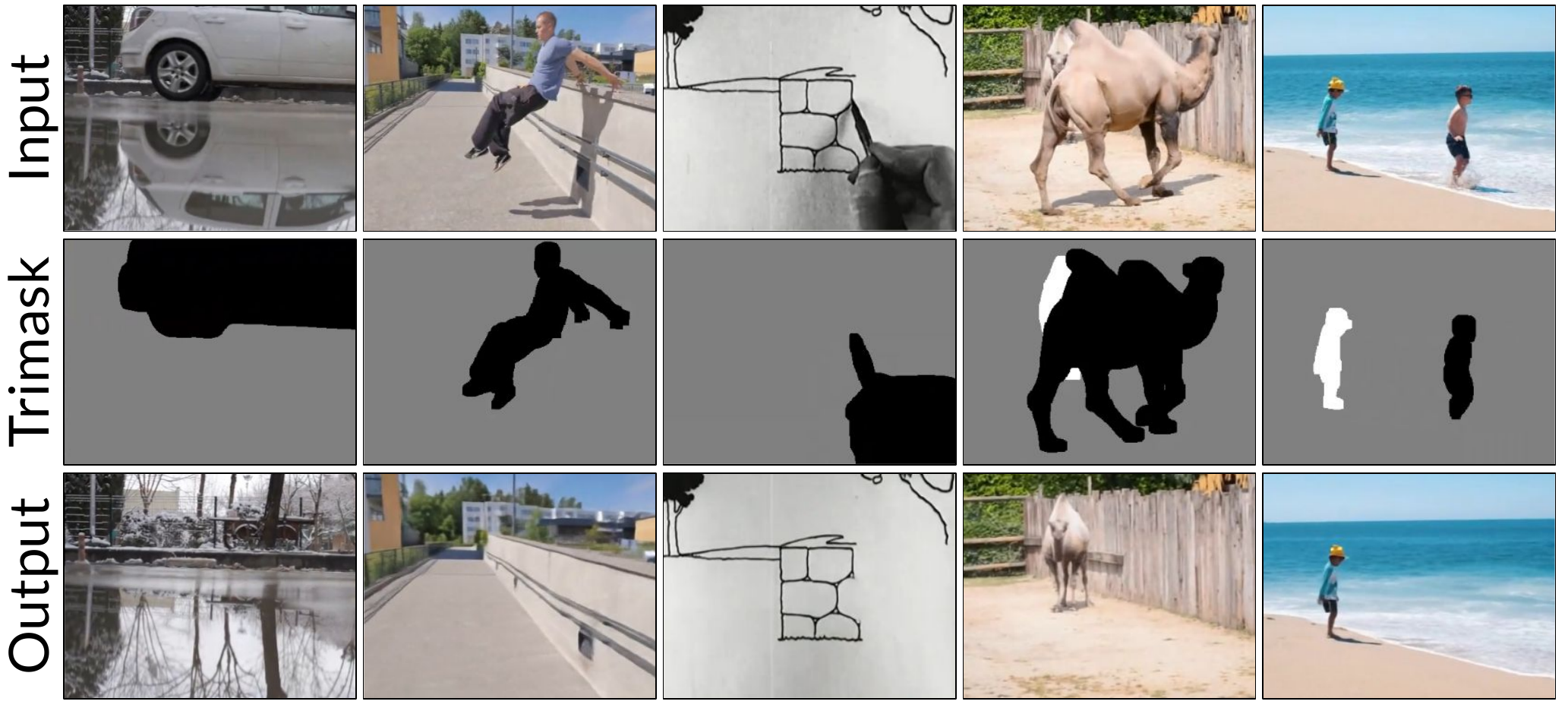}
\caption{\textbf{Object-effect-removal results of CogVideoX-based Casper.}}
\label{fig:public_casper}
\end{figure}

\paragraph{Reproducibility.}
We will release our dataset for reproducibility. In addition, we finetuned a publicly available CogVideoX~[56] using the same data to create a DiT version of Casper\footnote{For CogVideoX-based Casper finetuning, we adopt the codes and model from the third-party github: \url{https://github.com/aigc-apps/CogVideoX-Fun.git}}, without adjusting the original CogVideoX hyperparameters.
The results are shown in Fig.~\ref{fig:public_casper}. The sampling process takes 66 sec for an 85-frame, 384$\times$672 video and a trimask with 50 DDIM steps w/o CFG on an A100 GPU.

\begin{figure}
\centering
\includegraphics[trim={0cm 26cm 44.6cm 0},clip,width=\linewidth]{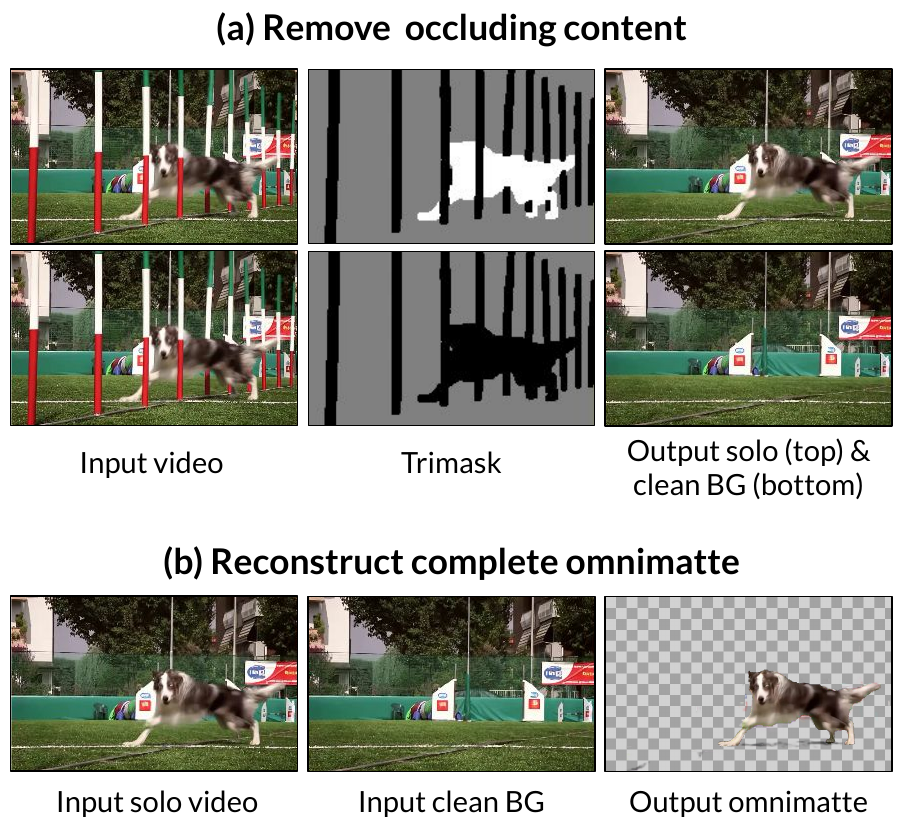}
\caption{\textbf{Handle occluding background content.}
Our method can also handle scenarios where the foreground object is occluded by background content (\eg, poles). By treating the occluding poles as additional foreground objects and removing them, we obtain a complete solo video of the dog and a clean-plate background video. These two videos can then be used to reconstruct the complete omnimatte of the dog.
}
\label{fig:supp_occlusion}
\end{figure}

\paragraph{Handling occluded foreground.}
Our Casper model can also handle certain occlusion scenarios where the foreground object is partially obscured by background content (Fig.~\ref{fig:supp_occlusion}). We can treat the occluding background content as additional foreground objects and remove them to reveal the fully visible target object.
Subsequently, the omnimatte optimization process utilizes the completed solo video and the clean background without occlusions as inputs to generate a complete omnimatte layer.

\begin{figure}
\centering
\includegraphics[trim={0cm 34cm 42.5cm 0.05cm},clip,width=0.97\linewidth]{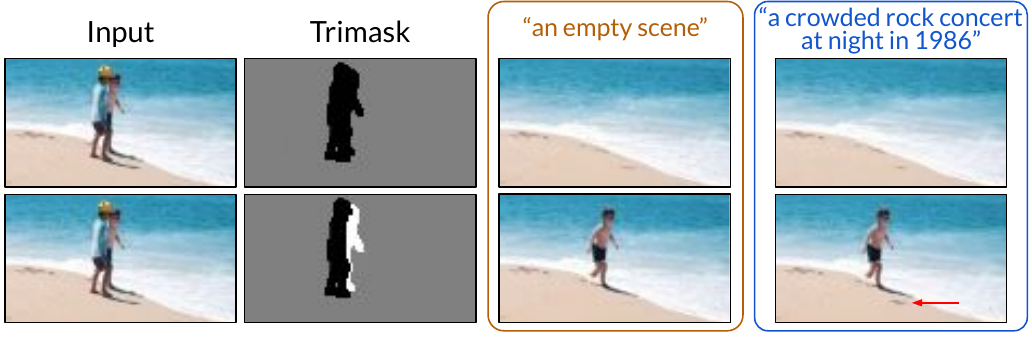}
\vspace{-0.3cm}
\caption{\textbf{Impact of the text prompt.}}
\label{fig:supp_text_prompts}
\end{figure}

\paragraph{Impact of text prompts.}
We used simple, short prompts such as ``a clean beach'' for fine-tuning the Lumiere-based Casper. 
While Casper is primarily driven by the input video and trimask and less sensitive to text prompts (``empty scene'' works well for both rows below), we did observe that a highly unrelated prompt (\eg, ``rock concert'') could affect the performance, as shown in Fig.~\ref{fig:supp_text_prompts}.


\paragraph{Challenges in handling similar objects.}
While our Kubric dataset includes multiple similar objects to help Casper handle such cases, it may still struggle with complex scenarios. We observed a similar issue in CogVideoX-based Casper, potentially due to the domain discrepancies between Kubric data and real-world videos. 
This could be mitigated in future work through more realistic data generation and re-introducing successful removal results into the training set.

\paragraph{Undesired detail changes.}
These artifacts are caused by Lumiere's SSR model, which can hallucinate high-frequency details. While we use a post-processing step to transfer original details, it is applied conservatively to avoid altering effect-removal areas. 
A future direction may be exploring an SSR model that can refer to the original video when upsampling the removal result.

\section{Training Details of Our Casper Model}
\label{sec:supp_training}

\paragraph{Omnimatte data.}
For the \emph{Omnimatte} training data,
we observed that the background video reconstructions produced by existing omnimatte methods~\cite{omnimatte,omnimatte3d,omnimatterf} can lack sharpness and alignment with the original input video.
This can lead to degradation of the original details in the preservation regions of our removal model, which is trained on pairs of original inputs and blurry background videos.
To address this issue,  we use the video reconstruction results of omnimatte methods as training inputs instead of the original videos. Since the video reconstruction is composed of omnimatte layers, it maintains the same quality as the target background video, preventing the model from learning quality degradation. 

\paragraph{Kubric data generation.}
To synthesize our \emph{Kubric} data, we randomly generate 1 to 6 objects in a scene. We also introduce challenging scenarios such as stationary objects, varying lighting conditions, and reflective surfaces.
The Kubric generation script will be publicly released.

\paragraph{Object-Paste data.} Due to the challenge of real data collection, Object-Paste is synthesized to enhance inpainting and background preservation, but not effect removal. Though not always reflected in metrics, we observed qualitative improvements in the main paper Fig.~9a and Supp. HTML. A tradeoff exists between background preservation and effect removal, and a high Object-Paste weight may hinder effect-removal performance. To balance this, we set its weight to 2\% in the training set.

\paragraph{Trimask label for synthetic data.}
For both synthetic \emph{Kubric} and \emph{Object-Paste} data, we randomly switch the labels from white to gray, to encourage the model to learn preservation and inpainting capabilities in gray-labeled background areas.

\paragraph{Data augmentation.}
Video examples are randomly clipped into 80-frame segments and cropped to a 128$\times$128px resolution for training. For real videos shorter than 80 frames, we apply temporal reflective padding to achieve the desired length.
We augment the four different categories of data with different ratios during the training.
Real-world examples from \emph{Omnimatte} and \emph{Tripod} constitute 50\% of the training data, while synthesized \emph{Kubric} and \emph{Object-Paste} data account for approximately 48\% and 2\%, respectively.

We fully finetune our Casper model from the pretrained Lumiere inpainting model~\cite{lumiere,vidpanos} for 20,000 iterations and a batch size of 32.

\section{Inference Details of Object and Effect Removal}
\label{sec:supp_inference}
Our Casper model takes 256 DDPM~\cite{ddpm} sampling steps without using classifier free guidance.
It is important to note that we do not cherry pick random seeds for object and effect removal outputs. We consistently apply a fixed random seed ($=0$) to all input videos.

\section{Omnimatte Optimization Details}
\label{sec:supp_omptimization}
The Casper model has a resolution of 128px (\eg, 224$\times$128) inherited from the Lumiere base stage.
While Lumiere's SSR stage~\cite{lumiere} upsamples the removal videos (\eg, $\vidbg$) to higher resolutions, (\eg, $\vidbg^\ssr$), it may hallucinate high-frequency detail in unpredictable ways.
Thus, directly reconstructing the omnimattes from the upsampled videos may result in noisy foreground layers, capturing unwanted background artifacts. 

To mitigate this issue, we employ a bootstrapping strategy.
Initially, we optimize the omnimatte with the base model outputs $(\vid_i, \vidbg)$ to obtain the alpha maps at the base 128px resolution.
We then use a bilinear upsampling of the 128px alpha map to supervise the optimization of a higher-resolution omnimatte (\eg, 640$\times$384) from the upsampled pair $(\vid^\ssr_i, \vidbg^\ssr)$ for finer details. 
The optimizations of base and upsampled resolutions use the same algorithm (Sec. 3.5 of main paper) but with slightly different hyper-parameter settings.

\begin{table*}
\caption{\textbf{Quantitative comparison.}
Following the benchmark established in OmnimatteRF~\cite{omnimatterf}, we evaluate the effect-removal quality on background videos of 10 synthetic scenes. Our method achieves the best overall scores in both PSNR and LPIPS metrics. We adopt the numbers of~\cite{omnimatte,layeredatlas,omnimatterf} reported in OmnimatteRF~\cite{omnimatterf}. Best results are highlighted in red and second-best in yellow. Results marked as ``-'' indicate failures (\eg, all zeros in Omnimatte3D~\cite{omnimatte3d}).
}
\vspace{-0.2cm}
\resizebox{\textwidth}{!}{
\begin{tabular}{l|cccccccccccccccccccc|cc}
\toprule
Scene & \multicolumn{2}{c}{Movie-Donkey} & \multicolumn{2}{c}{Movie-Dog} & \multicolumn{2}{c}{Movie-Chicken} & \multicolumn{2}{c}{Movie-Rooster} & \multicolumn{2}{c}{Movie-Dodge} & \multicolumn{2}{c}{Kubric-Car} & \multicolumn{2}{c}{Kubric-Cars} & \multicolumn{2}{c}{Kubric-Bag} & \multicolumn{2}{c}{Kubric-Chair} & \multicolumn{2}{c|}{Kubric-Pillow} & \multicolumn{2}{c}{Average} \\
Metric & PSNR$\uparrow$ & LPIPS$\downarrow$ & PSNR$\uparrow$ & LPIPS$\downarrow$ & PSNR$\uparrow$ & LPIPS$\downarrow$ & PSNR$\uparrow$ & LPIPS$\downarrow$ & PSNR$\uparrow$ & LPIPS$\downarrow$ & PSNR$\uparrow$ & LPIPS$\downarrow$ & PSNR$\uparrow$ & LPIPS$\downarrow$ & PSNR$\uparrow$ & LPIPS$\downarrow$ & PSNR$\uparrow$ & LPIPS$\downarrow$ & PSNR$\uparrow$ & LPIPS$\downarrow$ & PSNR$\uparrow$ & LPIPS$\downarrow$ \\ 
\midrule
ObjectDrop~\cite{objectdrop} & 27.23 & 0.091 & 28.84 & 0.129 & 27.87 & 0.153 & 26.68 & 0.145 & 29.64 & 0.102 & 33.92 & 0.087 & 34.31 & 0.101 & 32.13 & 0.051 & 34.95 & 0.081 & 35.80 & 0.096 & 31.14 & 0.104 \\
Propainter~\cite{propainter} & 27.09 & 0.133 & 27.67 & 0.109 & 26.82 & 0.119 & 24.31 & 0.143 & 31.31 & 0.065 & 32.59 & 0.075 & 34.54 & 0.081 & 32.52 & 0.046 & 34.99 & 0.047 & 38.72 & 0.030 & 31.06 & 0.085 \\
Lumiere inpainting~\cite{lumiere} & 25.31 & 0.157 & 26.97 & 0.159 & 26.60 & 0.162 & 24.39 & 0.163 & 29.82 & 0.101 & 30.05 & 0.201 & 31.04 & 0.202 & 29.32 & 0.165 & 32.11 & 0.143 & 34.77 & 0.075 & 29.04 & 0.153 \\
Ominmatte~\cite{omnimatte} & 19.11 & 0.315 & 21.74 & 0.279 & 20.95 & 0.312 & 23.14 & 0.220 & 23.88 & 0.067 & 31.14 & 0.162 & 31.20 & 0.157 & 23.64 & 0.271 & 26.91 & 0.175 & 21.17 & 0.270 & 24.29 & 0.223 \\
LNA~\cite{layeredatlas} & 18.79 & 0.104 & 26.08 & 0.154 & 19.22 & 0.190 & 26.46 & 0.131 & 24.94 & 0.068 & - & - & - & - & 27.08 & 0.138 & 21.21 & 0.105 & 31.66 & 0.080 & - & - \\ 
Omnimatte3D~\cite{omnimatte3d} & 24.72 & 0.234 & 23.15 & 0.372 & 24.17 & 0.266 & 23.98 & 0.372 & - & - & 34.61 & 0.142 & 36.48 & 0.126 & 33.94 & 0.135 & - & - & 37.01 & 0.119 & - & - \\ 
OmnimatteRF~\cite{omnimatterf} & \cellcolor[HTML]{FFCCC9}38.24 & \cellcolor[HTML]{FFCCC9}0.005 & \cellcolor[HTML]{FFEEBF}31.44 & \cellcolor[HTML]{FFCCC9}{\color[HTML]{333333} 0.030} & \cellcolor[HTML]{FFCCC9}32.86 & \cellcolor[HTML]{FFCCC9}0.021 & \cellcolor[HTML]{FFEEBF}27.65 & \cellcolor[HTML]{FFCCC9}0.024 & \cellcolor[HTML]{FFCCC9}39.11 & \cellcolor[HTML]{FFCCC9}0.006 & \cellcolor[HTML]{FFEEBF}39.09 & \cellcolor[HTML]{FFEEBF}0.033 & \cellcolor[HTML]{FFEEBF}39.78 & \cellcolor[HTML]{FFEEBF}0.032 & \cellcolor[HTML]{FFEEBF}39.58 & \cellcolor[HTML]{FFEEBF}0.029 & \cellcolor[HTML]{FFEEBF}42.46 & \cellcolor[HTML]{FFEEBF}0.023 & \cellcolor[HTML]{FFEEBF}43.62 & \cellcolor[HTML]{FFEEBF}0.022 & \cellcolor[HTML]{FFEEBF}37.38 & \cellcolor[HTML]{FFEEBF}0.023 \\
Ours & \cellcolor[HTML]{FFEEBF}32.02 & \cellcolor[HTML]{FFEEBF}0.017 & \cellcolor[HTML]{FFCCC9}33.33 & \cellcolor[HTML]{FFEEBF}0.033 & \cellcolor[HTML]{FFEEBF}32.59 & \cellcolor[HTML]{FFEEBF}0.037 & \cellcolor[HTML]{FFCCC9}29.31 & \cellcolor[HTML]{FFEEBF}0.047 & \cellcolor[HTML]{FFEEBF}36.20 & \cellcolor[HTML]{FFEEBF}0.014 & \cellcolor[HTML]{FFCCC9}42.78 & \cellcolor[HTML]{FFCCC9}0.011 & \cellcolor[HTML]{FFCCC9}44.41 & \cellcolor[HTML]{FFCCC9}0.016 & \cellcolor[HTML]{FFCCC9}42.96 & \cellcolor[HTML]{FFCCC9}0.007 & \cellcolor[HTML]{FFCCC9}43.94 & \cellcolor[HTML]{FFCCC9}0.011 & \cellcolor[HTML]{FFCCC9}46.25 & \cellcolor[HTML]{FFCCC9}0.006 & \cellcolor[HTML]{FFCCC9}38.38 & \cellcolor[HTML]{FFCCC9}0.020 \\
\bottomrule
\end{tabular}
}
\label{tbl:quantitative_comparison_per_scene}
\end{table*}

\begin{table}
\caption{\textbf{Ablation study on our training data.}
To assess the individual contribution of each data category, we conduct an ablation study by incrementally adding each category to the training set. We encourage the readers to view our HTML file for visual comparisons on in-the-wild videos.
}
\resizebox{
\linewidth}{!}{
\begin{tabular}{cccccc}
\toprule
\multicolumn{4}{c}{Training data category} & \multicolumn{2}{c}{Metric} \\
Omnimatte & Tripod & Kubric & Object-Paste & PSNR$\uparrow$ & LPIPS$\downarrow$ \\ \midrule
\checkmark & \xmark & \xmark & \xmark & 37.06 & 0.027 \\
\checkmark & \checkmark & \xmark & \xmark & 36.97 & 0.026 \\
\checkmark & \checkmark & \checkmark & \xmark & 38.36 & 0.020 \\
\checkmark & \checkmark & \checkmark & \checkmark & 38.38 & 0.020 \\ \bottomrule
\end{tabular}
}
\label{tbl:supp_ablation_data}
\end{table}

\begin{table}
\caption{\textbf{Ablation study on the input conditions for the Casper model.}
The original inpainting condition utilizes a binary mask, while the video condition involves zeroing out the removal region. Following ObjectDrop~\cite{objectdrop}, the content within the removal regions is preserved in the condition to enable the model to associate effects outside the mask with the content inside.
Finally, we replace the binary mask condition with our proposed trimask to mitigate ambiguity in effect removal within preservation regions. The full impact of these input conditions may not be evident from the evaluation of 10 synthetic background videos. We therefore encourage readers to examine comparisons on real-world videos in our supplementary HTML.
}
\centering
\resizebox{
\linewidth}{!}{
\begin{tabular}{lcccc}
\toprule
 & RGB video condition & Mask condition & PSNR$\uparrow$ & LPIPS$\downarrow$ \\ \midrule
Original inpainting & Masking removal area & Binary & 38.58 & 0.021 \\
ObjectDrop approach~\cite{objectdrop} & No masking & Binary & 38.24 & 0.020 \\
Our condition & No masking & Our trimask & 38.38 & 0.020 \\ \bottomrule
\end{tabular}
}
\label{tbl:supp_ablation_condition}
\end{table}
\topic{Base optimization}
We optimize the base resolution omnimattes using the following loss function: $\loss_{\mathrm{total}}=\loss_{\mathrm{recon}} + \lambda_\mathrm{sparsity}\loss_{\mathrm{sparsity}} + \lambda_\mathrm{mask}\loss_{\mathrm{mask}}$ to optimize the base resolution omnimattes.
The balancing weight $\lambda_{\mathrm{sparsity}}$ is set to 1, along with the constant weights $\beta_0$ and $\beta_1$ for the sparsity loss (Eq.~4 of the main paper) are empirically set to 1 and 10, respectively. 
The weight of mask supervision $\lambda_{\mathrm{mask}}$ is initialized to 20 and gradually reduced over the optimization.
The optimization takes 20,000 iterations with a batch size of 20 frames at a 128px base resolution of 128px.

\topic{Upsampling optimization}
To bootstrap higher resolution of omnimattes (\ie, foreground RGB $\vidfgi^{\mathrm{hr}}$ and alpha $\a^{\mathrm{hr}}_{i}$), we employ the solo video and background video of the SSR outputs, $(\vid_i^{\ssr}, \vidbg^{\ssr})$ as the input pair for our optimization framework.
The foreground RGB variables are initialized using the upsampled base-resolution optimized RGB $\vidfgi$, and an additional alpha supervision loss, $\loss_{\mathrm{alpha}}=\|\a^{\mathrm{hr}}_i - \a^{\mathrm{up}}_i\|_2$, is introduced, supervised by the upsampled base-resolution alpha maps, $\a^{\mathrm{up}}_i$.
To prevent the model from learning aliased boundaries, we disable supervision loss on the edge regions of the alpha maps. Moreover, we switch the photometric reconstruction loss, $\loss_{\mathrm{recon}}$ (Eq. 3 of the main paper), from L2 loss to L1 loss to mitigate the impact of outlier hallucinated high-frequency details produced by the SSR model.
The loss weights $\lambda_{\mathrm{sparsity}}$ and $\lambda_{\mathrm{mask}}$ are both initialized to 10, while the mask supervision loss $\loss_{\mathrm{mask}}$ is deactivated after the first 2,000 iterations.
The weight of the base-resolution alpha supervision loss $\lambda_{\mathrm{alpha}}$ is set to 20.
The optimization process runs for 20,000 iterations to obtain the final omnimattes.

\topic{Input video reconstruction and Detail Transfer}
To reconstruct the original input video from individual optimized omnimatte layers, depth information is required to determine the correct layer order in multi-object scenarios.
We utilize DepthCrafter~\cite{depthcrafter} to estimate video depth and define the frame-level depth order for foreground layers. Subsequently, all layers, including the clean background, are composited from back to front using the over operation~\cite{porter1984compositing}.

During the compositing process, we compute the composited opacity for each layer.
For layer pixels where the composited opacity reaches 1 (\ie, fully opaque), a detail transfer step~\cite{layeredneuralrendering,omnimatte,omnimatte3d} is applied.
This step copies the original details from the input video to the high-resolution omnimatte and background layers, mitigating misaligned high-frequency details that may have been hallucinated by the SSR model.

\section{Runtime}
\label{sec:supp_runtime}
For Stage 1, Casper takes 12 min to process an 80-frame video and 15-20 min for longer videos, the same as the original Lumiere-Inpainting. 
We run Casper on a 96GB TPU with a batch size of 4 (\eg, three solo videos and a clean background). 
Additional objects can be run in parallel with multiple TPUs.
For Lumiere SSR upsampling, it takes around 15 minutes.
For Stage 2, our unoptimized code takes 7 min to produce each object layer on a 48GB TPU.
Each object layer is computed independently and thus can be parallelized.
Once obtaining all layers, the post-processing detail transfer takes 1 minute.
For an 80-frame video of 3 objects on a single TPU, the entire process takes $12 + 15 + 7\times3 + 1$ = 49 min (or $12+15+7+1=35$ min with multi TPUs).
In contrast, OmnimatteRF takes 3 hr to optimize a NeRF and all layers together, potentially limiting the number of objects due to GPU memory constraints.

\section{Onimatte-RF Synthetic Evaluation Benchmark}
\label{sec:supp_eval}

Table~\ref{tbl:quantitative_comparison_per_scene} presents the per-scene evaluation scores. Omnimatte3D~\cite{omnimatte3d} fails to reconstruct the background layer in two scenes, resulting in all-zero outputs. OmnimatteRF~\cite{omnimatterf} employs an additional background retraining step to enhance effect removal and inpainting accuracy in certain cases by leveraging a global background scene model. Our method performs the overall best in both PSNR and LPIPS~\cite{lpips} metrics.

\section{Ablation study}
\label{sec:supp_ablation}
\topic{Training data.}
Table~\ref{tbl:supp_ablation_data} presents a quantitative ablation study on the 10 synthetic evaluation scenes from OmnimatteRF~\cite{omnimatterf}. By incrementally adding four distinct data categories to the training set, the removal model achieves improved effect removal performance. For further video comparisons on real-world videos, please refer to our supplementary HTML.

\topic{Input condition.}
Table~\ref{tbl:supp_ablation_condition} illustrates the quantitative performance of removal models trained with various input condition settings. The original inpainting condition masks out the RGB values within the removal regions and concatenates them with a binary mask. Following ObjectDrop~\cite{objectdrop}, we unmask the RGB values in the removal regions to enable the model to associate effects with content. Finally, we introduce our proposed trimask to replace the binary mask condition, alleviating ambiguity in effect removal within preservation regions. While the effectiveness of these input conditions may not be readily apparent from the evaluation of 10 synthetic background videos, we encourage readers to examine comparisons on real-world videos in our supplementary HTML.

\end{document}